\def\year{2022}\relax
\documentclass[letterpaper]{article} 
\usepackage{aaai22}  
\usepackage{times}  
\usepackage{helvet}  
\usepackage{courier}  
\usepackage[hyphens]{url}  
\usepackage{graphicx} 
\usepackage{natbib}  
\usepackage{caption} 
\DeclareCaptionStyle{ruled}{labelfont=normalfont,labelsep=colon,strut=off} 
\usepackage{algorithm}
\usepackage{algorithmic}

\usepackage{xcolor}

\usepackage{color}

\usepackage[caption=false]{subfig}

\usepackage{newfloat}
\usepackage{listings}
\floatstyle{ruled}
\newfloat{listing}{tb}{lst}{}
\floatname{listing}{Listing}
\copyrighttext{Presented at the AI-HRI Symposium at AAAI Fall Symposium Series (FSS) 2022}

\title{Taxonomy of A Decision Support System for Adaptive Experimental Design in Field Robotics}
\author{
    Jason M. Gregory\textsuperscript{\rm 1}, Sarah Al-Hussaini\textsuperscript{\rm 2}, Ali-akbar Agha-mohammadi\textsuperscript{\rm 3}, Satyandra K. Gupta\textsuperscript{\rm 2}
}
\affiliations{
    \textsuperscript{\rm 1} DEVCOM Army Research Laboratory, Adelphi, MD, USA\\
    \textsuperscript{\rm 2} Viterbi School of Engineering,
    University of Southern California, CA, USA\\ 
    \textsuperscript{\rm 3} NASA Jet Propulsion Laboratory, Pasadena, CA, USA\\
    jason.m.gregory1.civ@army.mil, salhussa@usc.edu, aliakbar.aghamohammadi@jpl.nasa.gov, guptask@usc.edu
}

\begin{document}

\maketitle

\begin{abstract}

Experimental design in field robotics is an adaptive human-in-the-loop decision-making process in which an experimenter learns about system performance and limitations through interactions with a robot in the form of constructed experiments.
This can be challenging because of system complexity, the need to operate in unstructured environments, and the competing objectives of maximizing information gain while simultaneously minimizing experimental costs.
Based on the successes in other domains, we propose the use of a Decision Support System (DSS) to amplify the human's decision-making abilities, overcome their inherent shortcomings, and enable principled decision-making in field experiments.
In this work, we propose common terminology and a six-stage taxonomy of DSSs specifically for adaptive experimental design of more informative tests and reduced experimental costs.
We construct and present our taxonomy using examples and trends from DSS literature, including works involving artificial intelligence and Intelligent DSSs.
Finally, we identify critical technical gaps and opportunities for future research to direct the scientific community in the pursuit of next-generation DSSs for experimental design.


\end{abstract}


\section{Introduction}
\label{sec:intro}

Robots that operate in harsh environments, i.e., field robots \cite{thorpe2001field}, typically consist of very complex systems and require experimentation in real-world settings to fully characterize behaviors, capabilities, and limitations. 
This is because the natural world is unstructured and introduces spatio-temporal effects, non-negligible stochasticity, and sources of system failure.
Importantly, a human is required to lead field robot experimentation in order to effectively prioritize objectives so testing is consistent with the scope of the specific field robotic application, manage risk to personnel and hardware, and provide context-dependent reasoning for desired system behaviors.
The construction of experiments, referred to here as \textit{experimental design}, must be carefully considered to maximize the information gained by the experimenter while simultaneously minimizing the cost (e.g., money, energy, and time to implement, set up, and execute) and risk (e.g., personal injury, system damage, and wasted resources).
By observing humans conducting field experiments, our previous work discovered that experimental design consists of a sequential decision-making process~\cite{gregory2022}.
A human selects a specific system configuration to test (e.g., hardware, software components, and parameter values) based on their hypotheses and then uses new information collected from empirical observations of the previous experiment to inform the decision for the next experiment.
During this adaptive, human-centric process, the experimenter effectively learns about the robotic system through interactions in the form of experiments.
Therefore, we view adaptive experimental design as a unique problem within the field of human-robot interaction (HRI). 

Researchers in the human-machine interaction community have found that a \textit{Decision Support System} (DSS) can vastly improve a human's decision quality when interacting with very complex systems \cite{ginzberg1982decision,aronson2005decision,antony2014design,bonczek2014foundations,anderson2018design}.
DSSs have been well studied for several decades \cite{eom1990survey,eom1998survey,eom2006survey,merkert2015survey} and have demonstrated significant impact in a number of different domains, including managerial~\cite{power2002decision}, medical~\cite{satapathy2022improved}, forestry~\cite{wikstrom2011heureka}, architectural~\cite{gercek2019energy}, warehouse management~\cite{accorsi2014decision}, and housing evaluation~\cite{natividade2007multicriteria} problem settings.

Specific to the robotics community, DSSs have been combined with robots for ocean exploration~\cite{gomes2013odss}, construction processes~\cite{marcher2021design}, and underwater sampling~ \cite{zhou2018adaptive}.
Researchers have also investigated incorporating artificial intelligence (AI) in a DSS, often times referred to as a \textit{Intelligent DSS} (IDSS)~\cite{kaklauskas2015intelligent,aronson2005decision,tariq2012intelligent}, for problems such as robotic surface vehicle navigation~\cite{shishkin2019intelligent} and threat assessment~\cite{phillips2002agent}.
DSSs and IDSSs in HRI have been used primarily in literature to improve robot decision making to perform some task or human decision making in the physical design of robotic systems~\cite{heikkila2013decision}, i.e., not in experimental design to explore system performance.
Because machines excel at reasoning over large-scale data, estimating uncertainty, and making unbiased decisions, we believe that there is a great opportunity to develop DSSs and IDSSs for the experimental design of field robotic systems. 

\begin{figure*}[ht!]
	\centering
  \subfloat[Stage $1$: Design Assistance \label{subfig:stage1-diagram}]{%
        \includegraphics[width=0.51\linewidth]{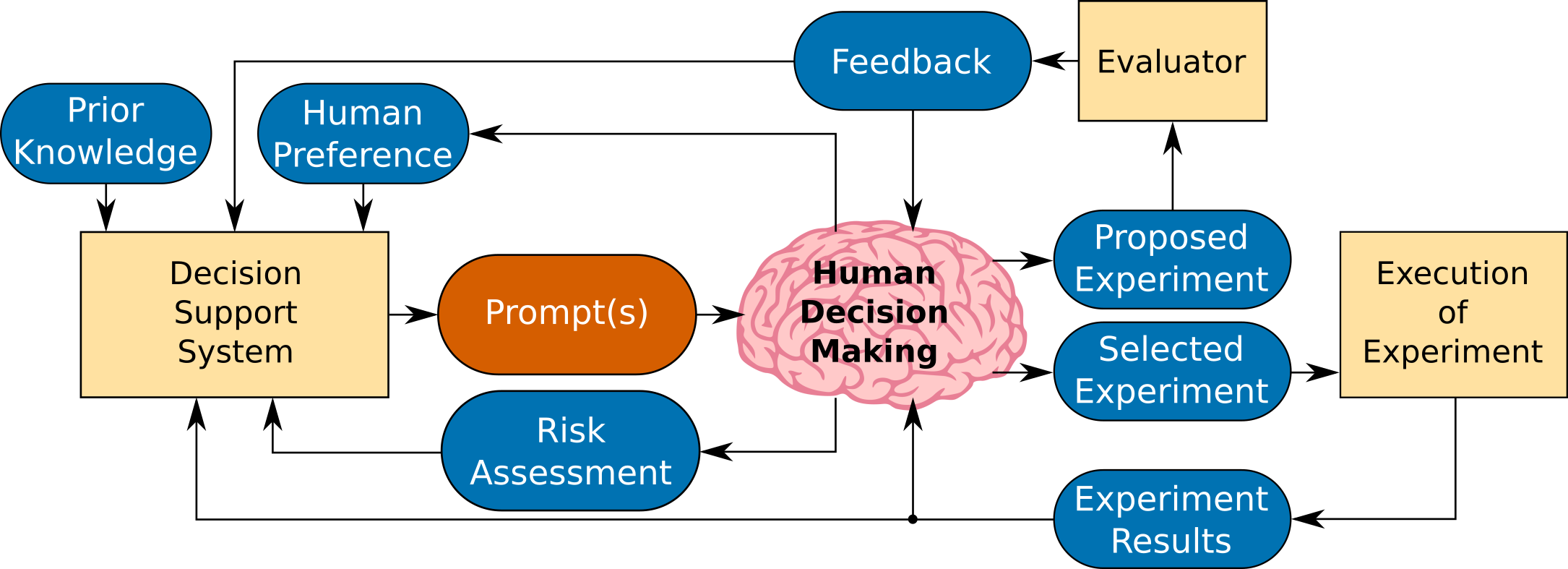}}

  \subfloat[Stage $2$: Design Monitoring \label{subfig:stage2-diagram}]{%
        \includegraphics[width=0.51\linewidth]{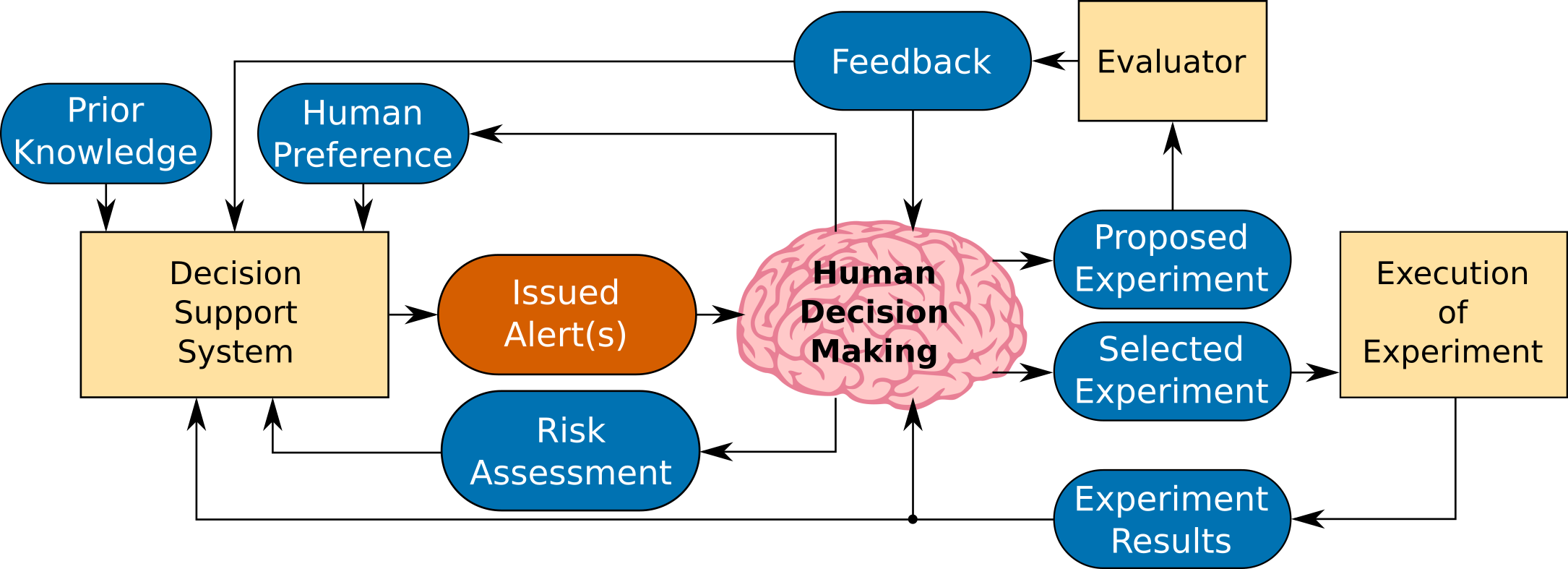}}

  \subfloat[Stage $3$: Conditional Design Recommendation \label{subfig:stage3-diagram}]{%
        \includegraphics[width=0.51\linewidth]{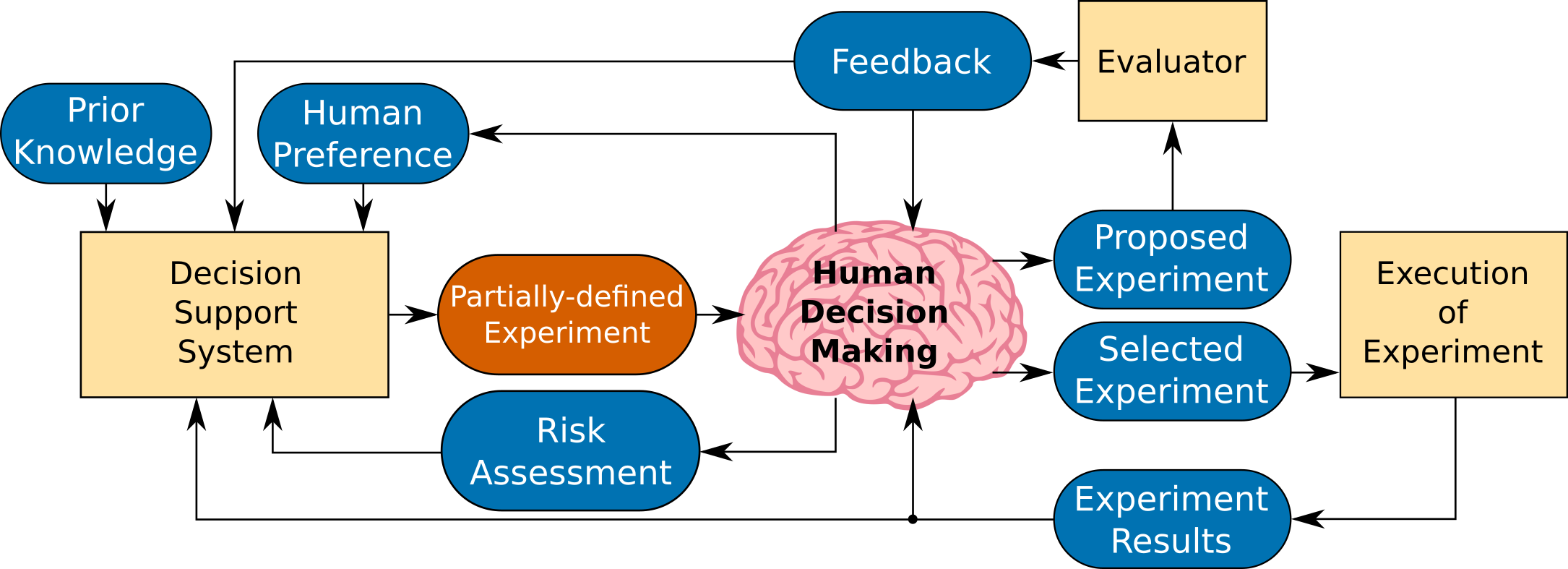}}

  \subfloat[Stage $4$: Single Design Recommendation \label{subfig:stage4-diagram}]{%
        \includegraphics[width=0.51\linewidth]{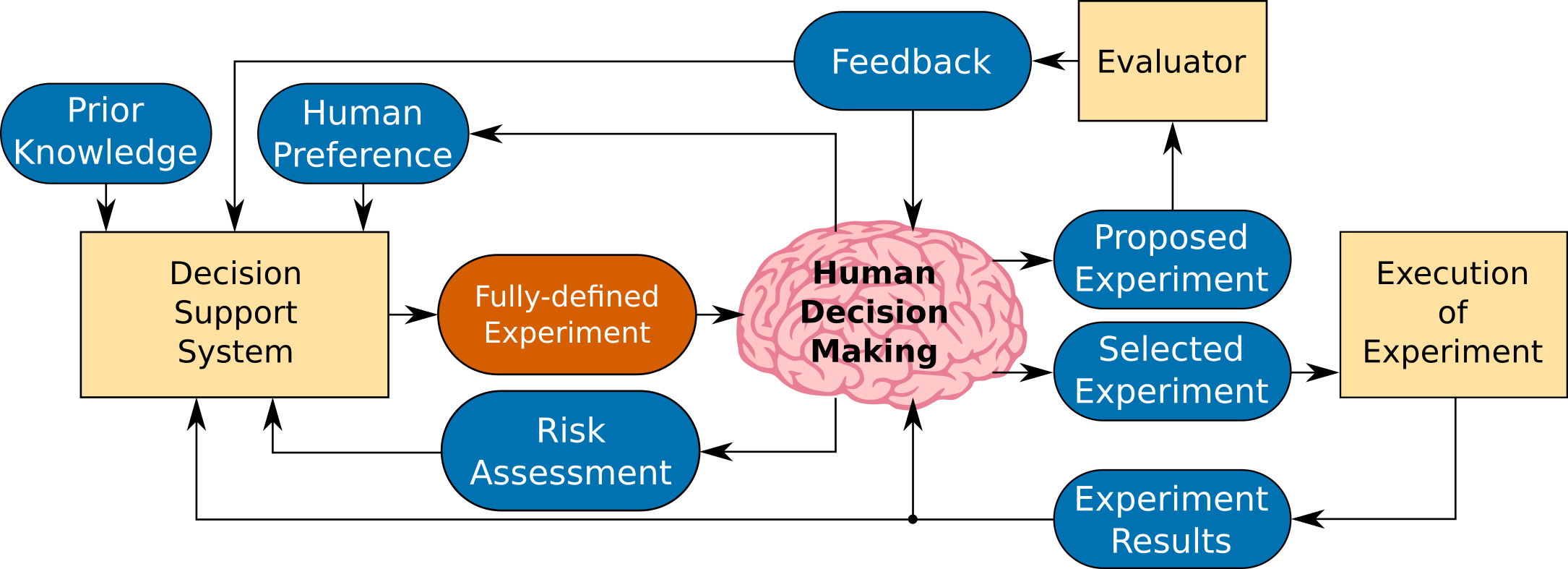}}

  \subfloat[Stage $5$: Sequential Design Recommendation \label{subfig:stage5-diagram}]{%
        \includegraphics[width=0.51\linewidth]{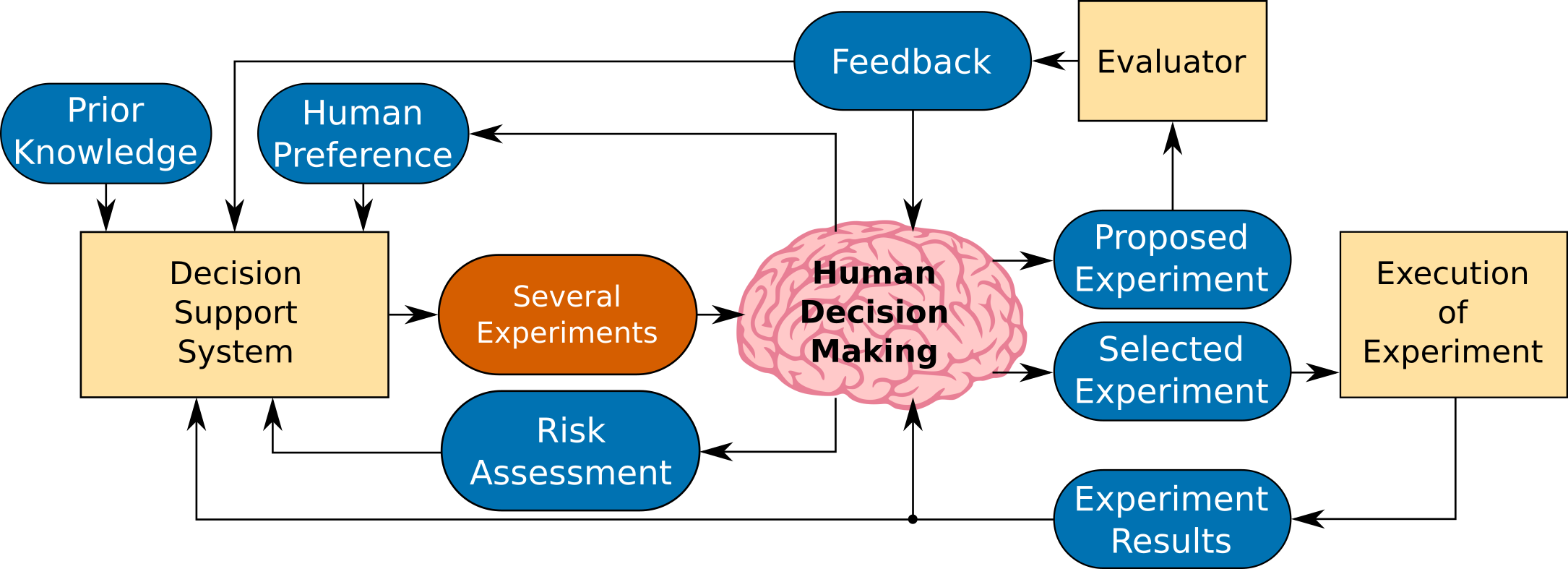}}
  	\caption{The different stages of a \textit{Decision Support System} (DSS) for adaptive experimental design where there is significant human-robot interaction. Stage $0$ (not shown) consists of a human-only decision making cycle where there is no experimental design decision support. An evaluator provides feedback to both the experimenter and, for stages $1$--$5$, the DSS regarding the proposed experiment from which both agents can learn and make new considerations before the human ultimately selects an experiment to conduct. }
  	\label{fig:dss-diagrams}
\end{figure*}

Given the absence of IDSSs for experimental design in HRI, and inspired by the large body of DSS literature in other domains, we seek to build a taxonomy to guide the research and development of IDSSs. 
The goal of this work is to lay the groundwork for defining terminology and functionality requirements that the scientific community can collectively investigate in the coming years. 
We envision DSSs, conceptually depicted in Figure~\ref{fig:dss-diagrams} and outlined in the Stages of Decision Support System Section, that can lead to the selection of more informative tests and subsequently reduce experimental costs by providing different forms of decision support to the human experimenter.
Ultimately, the experimenter is the decision maker with full authority over which experiment is conducted, including the ability to partially accept or reject support from a DSS, but we believe such an advisory system can greatly improve the quality of the human's decisions.
%

\textbf{Contributions.}
In this work, we: 
\begin{enumerate}
    \item define common terminology for adaptive experimental design of field robotic systems to facilitate more coordinated research efforts by the scientific community.
    \item provide an overview of literature regarding existing works of DSS and IDSS development and an analysis of general themes and requirements.
    \item propose a six-stage taxonomy of DSSs and IDSSs for adaptive experimental design in field robotics informed by systems presented in the literature. We believe that this roadmap provides the scientific community with common goals and requirements that are necessary to realize principled experimental design solutions. We define each stage, the experimenter's responsibilities, and the feature requirements for the DSS. 
    \item identify critical technical gaps with respect to our proposed taxonomy to help guide future research toward the realization of different DSS solutions.
\end{enumerate}
\section{Terminology} \label{sec:terminology}

In this section we define necessary terms in the context of adaptive experimental design for field robotics.
This seeks to provide a common terminology for researchers and reduce confusion given that existing DSS development spans many different domains.
\begin{itemize}
    \item \textit{Autonomy Under Test (AUT)}: a reconfigurable field robotic system, equipped with autonomous capabilities, that an experimenter seeks to understand through experimentation. 
    \item \textit{Study}: a series of one or more iterations of the scientific process, including hypothesis construction, experimental design, conducting a field experiment, and gathering and analyzing results.
    \item \textit{Experimentation}: the process of testing hypotheses about the AUT. For clarity, automating experimentation is not within the scope of this taxonomy.
    \item \textit{Experimenter}: the human decision maker that chooses and conducts the field experiments.
    \item \textit{Experiment}: one iteration of evaluating the AUT with respect to selected inputs in order to obtain an observation. 
    \item \textit{Experimental design}: the construction of field experiments, which consists of defining experiment inputs. Proposing a taxonomy for DSSs that assist with experimental design is the focus of this paper.  
    \item \textit{Input}: a fully-defined configuration of the subsystems, components, and parameters that comprise a field experiment.
    \item \textit{Observation}: a record related to the outcome of an experiment, including both quantitative and qualitative performance.
    \item \textit{Decision Support System (DSS)}: a system that aids the experimenter with making decisions pertaining to experimental design.
    \item \textit{Subsystem}: a self-contained system within the AUT comprised of one or more components (e.g., a perception subsystem or a motion planning subsystem).
    \item \textit{Component}: a reconfigurable class of elements in an AUT's subsystem (e.g., hardware on the AUT such as a perception sensor or software capabilities used by the AUT such as a global motion planner).
    \item \textit{Parameter}: a reconfigurable characteristic in a set that collectively defines the configuration of a component (e.g., a specific LiDAR sensor or value for replanning rate) or experimental control variable (e.g., location of navigation goals).
\end{itemize}

\section{Related Works}
\label{sec:related}

In this section, we provide an overview of the DSS and IDSS literature from a number of different domains.
We present related works categorized by operations, assistance type, and evaluation and metrics as way to capture general trends and themes in literature with respect to system structure.
We use the literature from these different categories to guide the development of our DSS taxonomy.

\subsection {Operations}

Research in DSSs dates back to the 1970s~\cite{alter1977taxonomy,keen1987decision,shim2002past}, originating in managerial decision support~\cite{little1979decision,ginzberg1982decision}, but has since spanned a vast range of other domains~\cite{phillips2017intelligent}.
Most often, DSSs operate with respect to a domain- or task-specific decision-making cycle.
For example, the work of \cite{povoroznyuk2019formalization} formalizes the stages of diagnostic and therapeutic measures in medical DSSs. 
Medical professionals can receive diagnostic information from DSSs in the form of prompts and alerts. 
These types of DSSs are being used widely and have reached acceptable performance for several medical conditions.
Future DSSs in the medical domain will seek to provide support for more challenging tasks, such as therapeutic measures in the form of treatment and rehabilitation recommendations.

DSSs also operate with respect to the time horizon in which decisions must be planned or made.
The authors of \cite{zhai2020decision} survey DSSs for agriculture and describe short-term planning (day-to-day activity), mid-term planning (seasonal), and long-term planning (yearly) for decision making. 
Their work highlights that the current agricultural DSSs mainly focus on short-term planning and lack considerations on mid-term and long-term planning. 
DSSs that account for longer term planning pose more technological challenges, but can provide more informed assistance towards better decision making. 

The taxonomy proposed in this paper defines the requirements for DSSs specifically for assisting an experimenter with the task of experimental design in the domain of field robotics.
We envision DSSs developed in the context of this taxonomy will operate in different stages (described in the Stages of Decision Support System Section), and the decision planning horizon will grow for higher stages.


\subsection {DSS Assistance Types}

In the literature, there are several forms of assistance that DSSs typically provide to the decision maker. 
One type of decision support can be viewed as \textit{information aiding decision making} and examples of this include alarms, ``what-if" analysis, and anomaly monitoring and detection \cite{van1998improving, kamis2008using,attia2012simulation,wiczorek2014benefits,shishkin2019intelligent,marcher2021design,al2021simulation,al2022seeking}. 
These DSSs provide some data to assist in the decision making process, which can result in improved decision making; however, much of the burden for decision making still lies with the human. 

A second, more sophisticated type of support is \textit{suggestion aiding decision making}, which typically take the form of recommendations of optimal decisions or decision parameters
\cite{salama2018decision,ghodsypour1998decision,hassan2017decision,liu2021clinical,udokporo2021decision,povoroznyuk2019formalization,ruiz2004decision}. 
DSSs that provide this type of assistance alleviate some of the decision-making burden by proactively suggesting decisions.
There are also works in literature that provide a combination of both information and suggestions for aiding decision making \cite{vasilyev2020development,gomes2013odss,tintarev2015explaining}.

The taxonomy presented in this paper spans multiple forms of decision support, where each stage offers unique assistance.
Stages $1$ and $2$ offer different amounts of decision support information, while stages $3$--$5$ provide experimental design recommendations of varying detail and time horizons.
Literature has also investigated human preference in decision support because it is typically beneficial to provide appropriate assistance catered to a specific decision maker \cite{schafer2002meta,marcher2021design,gomes2013odss}. 
Using this insight, the proposed taxonomy allows the experimenter to select the DSS stage for every experiment in a study so that the decision maker receives their preferred decision support.



\subsection {Evaluation and Metrics for DSS}

The evaluation process and metrics for DSSs is a well-studied topic in the literature.
In most cases, evaluation refers to \textit{ex-post} where the observed result of DSS implementation is evaluated, that is, the quality of the decision \cite{adelman1991experiments}. 
Examples of such evaluation are common in the medical \cite{montgomery2000evaluation} and marketing \cite{van1996impact} domains.  

Alternatively, literature has considered \textit{ex-ante} evaluation where either the impact of the DSS in the decision making process or the processes within the DSS are assessed.
This type of evaluation can help determine the information necessary to optimally design a DSS or optimally select a DSS portfolio \cite{te1991feedback,gardner1993design}. 
For example, the authors of \cite{riedel1986utilization} investigate how to determine which evaluation information to obtain to maximize utilization. 
The work of \cite{zhai2020decision} discusses evaluating components in agricultural DSS processes, such as prediction, expert knowledge, and use of historical data. 
Finally, some evaluations include process- and outcome-oriented evaluation \cite{phillips2004multiple}, as well as a combination of both \cite{shani2011evaluating}.

Dimensions and metrics for evaluating recommender systems is an active area of research in the literature. 
The authors of \cite{avazpour2014dimensions} present a range of metrics categorized by recommendation-, user-, system-, and delivery-centric evaluation.
Recommendation-centric evaluation includes measuring the correctness of recommendations, coverage, diversity, and confidence in the recommendation.
User-centered evaluation is defined by trustworthiness, serendipity, utility, and risk metrics. 
System-centric evaluation can be based on robustness, learning rate, scalability, and stability. 
Delivery-centric evaluation uses metrics such as usability and user preference. 


The six-stage taxonomy presented in this paper defines both the requirements of the DSS as well as the experimenter's responsibility for each stage of operation.
These are developed and presented to enable both ex-post and ex-ante evaluation as to well as leverage existing evaluation metrics from the literature. 
\section{Stages of Decision Support Systems}  \label{sec:dss-stages}
In this section, we propose a six-stage taxonomy of DSSs for adaptive, experimental design.
This taxonomy is inspired by literature and other domains, but is tailored to the unique challenges present in field robotics relating to complex systems and real-world environmental constraints. 
We begin by providing an overview of DSSs and then define the responsibilities of the experimenter and the requirements for each stage of the DSS.
%
%
\subsection{Concept of Operations} \label{subsec:conops}
A DSS for experimental design must first build a knowledge database using data from the experimenter and experiments in order to provide relevant decision support.
As shown in Figure~\ref{fig:dss-diagrams}, the DSS receives as input prior knowledge, the experimenter's preferences and risk tolerance, intermediate feedback from the evaluator, and all experimental results.
All of these sources of data could be inputted to the DSS manually (e.g., the experimenter verbally or textually specifies their own prior knowledge, risk tolerance, preference, or qualitative observations), but there are vast opportunities for both inputting and processing this data in an automated fashion in order to improve efficiency.
For example, a database could operate on multiple studies to curate prior knowledge, human preference and risk may be inferred by decision monitoring, and experimental results can be empirically monitored and analyzed.
We also envision a workflow where DSSs observe some number of experiments before providing any decision support in order to properly initialize a sufficiently rich database; this initialization process may also be dependent on the size, quality, and relevancy of prior knowledge applicable to the AUT and experimenter.

\begin{figure}[!h]
  \centering
  \includegraphics[width=0.90\columnwidth]{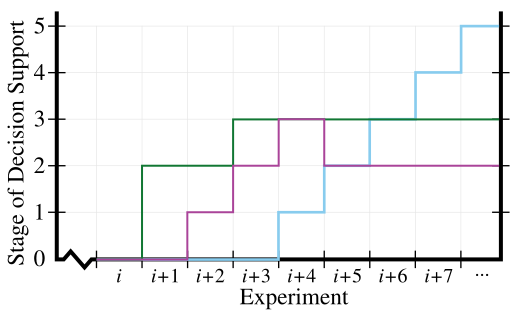}
\caption{ Three illustrative scenarios of adaptive experimental design in a hypothetical study containing several experiments. An experimenter can choose to monotonically increase the stages of decision support (blue line), increase until a certain stage but then plateau (green line), or increase and then decrease the DSS stage (purple line). Importantly, the experimenter can choose to use any stage (provided that the DSS has the necessary features) for any number of experiments and change the stage in response to new observations. }
  \label{fig:progression}
\end{figure}

\begin{figure*}[!htbp]
  \centering
  \includegraphics[width=2.1\columnwidth]{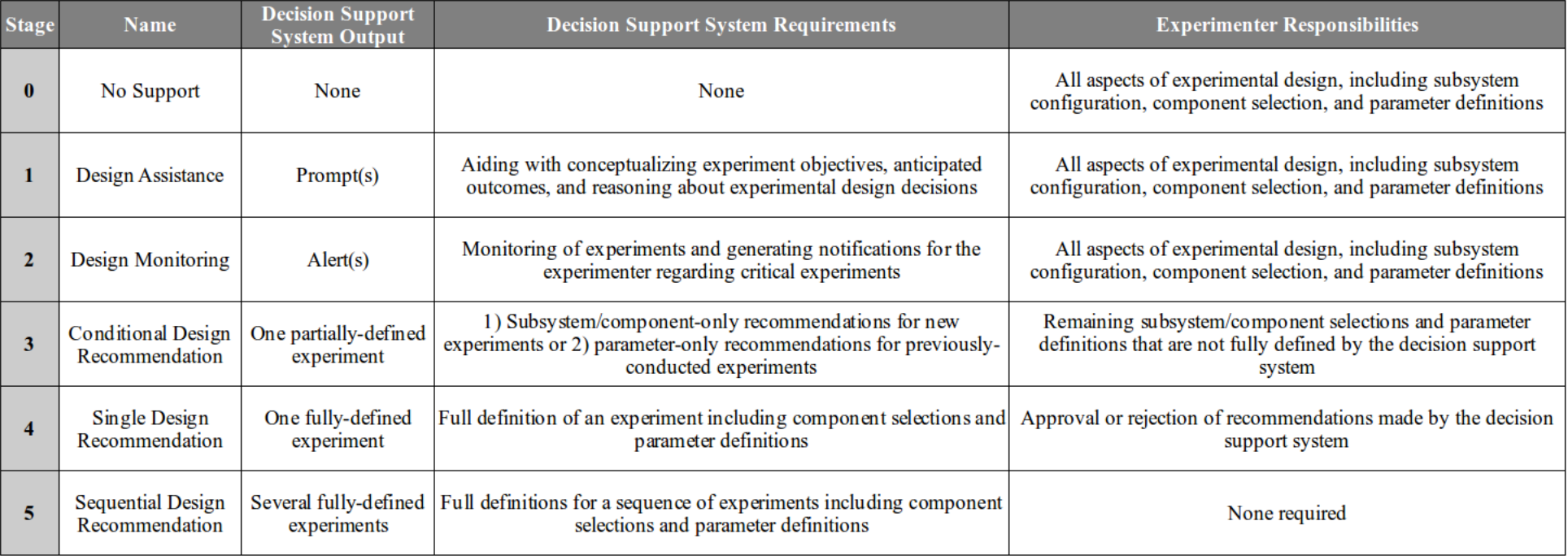}
  \caption{ An overview of the proposed six-stage taxonomy of design support systems for adaptive experimental design in field robotics. }
  \label{fig:stages}
\end{figure*}

We define the experimental design DSS operations in terms of \textit{stages}, where the experimenter selects one for each experiment in a study. 
A stage is defined by the functions and roles of the support system and are inversely proportional to the amount of experimenter's responsibilities.
As stages increase, the DSS inherits the functionality of the previous stage(s) and adds more automated capabilities, which, in turn, reduces the required responsibilities of the experimenter.
An illustrative example of this concept is presented in Figure~\ref{fig:progression} and an overview of each stage is summarized in Figure~\ref{fig:stages}.

In defining the requirements for each stage we draw inspiration from the levels of driving automation \cite{sae2018taxonomy}; however, adaptive experimental design is dependent on a fluid decision making process, which is an important distinction from self-driving vehicles.
To this end, the stage selection for a given experiment is the experimenter's decision and, unique to adaptive experimental design, can change throughout a series of experiments as the human performs more testing, obtains more data, and builds an understanding of the AUT within the experimenter-defined context.
Generic, information-based thresholds are intentionally not used to determine when to transition to different stages because the goal of experimentation is to build the experimenter's understanding of the AUT.
Defining machine-interpretable thresholds for generalizable, autonomous stage transitioning is likely intractable and also offers less control and flexibility to the experimenter than if they were to choose their desired decision support.
Humans play a central role because they determine if the DSS should graduate to any higher stages, to which stages the DSS should increase (given a sufficiently capable DSS) and when experiments should halt. 
Note that an experimenter is not required to increase the stage for any specific experiment, monotonically increase stages, or adjust stages incrementally.
We anticipate that experimenters will adjust the DSS stage to meet their needs, manage their workload, and transfer decision autonomy of experimental design analogous to the principles in the field of shared autonomy \cite{fridman2018human,schilling2016towards}.

Although not yet realized, we believe that continuous advances in AI and machine learning (ML) could serve as key enabling technologies for more sophisticated DSSs such that any stage in the taxonomy could take the form of an IDSS.
We envision that AI / ML-enabled DSSs will offer a range of capabilities, including searching the design space and alleviating the experimenter’s burden of identifying critical components and parameters, making actionable recommendations that the human might not otherwise consider, learning about the human’s preferences and risk tolerance through intuitive interactions to make recommendations more valuable to the experimenter, reminding the experimenter of vital experimental considerations to reduce the frequency of oversights and poor decisions, learning from the experimenter's mistakes just as a human would, adapting to changing conditions and new information also as a human would, and enabling intelligent selection of what data the DSS and the experimenter should learn from for more efficient experiments.
%
%

%

\subsection{Stage 0: No Support} \label{subsec:stage0}
In the first stage, stage $0$, the experimenter is responsible for all aspects of experimental design, including all component and parameter selections. 
Under these conditions, the DSS does not offer any information or recommendation, and the full burden of experimental design is placed on the experimenter.
Note that much of the experimental design in the current field robotics literature corresponds to stage $0$ in that there is no DSS or experiment recommendation.

\subsection{Stage 1: Design Assistance} \label{subsec:stage1}
A stage $1$ experimental design DSS assists the experimenter with conceptually formulating the experiments, as depicted in Figure~\ref{subfig:stage1-diagram}.
Similar to stage $0$, the DSS does not make any recommendations, but it does assist with experimental design by prompting the experimenter for considerations, predictions, and justifications of design decisions as a means for increasing the probability that experiments produce meaningful, defensible, and intended outcomes with respect to a specific hypothesis.
Field robotic systems and the environments in which they operate in are inherently complex, which motivates the use of simple prompts as a way to tractably support understanding and pursuing experiment objectives.
This stage is inspired by ``what-if" analysis used in other applications~\cite{van1998improving,kamis2008using} and could be implemented by different forms of interaction, including pen-and-paper, web-based tools, or verbal communication.
Furthermore, as in stage $0$, the experimenter is responsible for all decisions pertaining to the experimental design. 

\subsection{Stage 2: Design Monitoring} \label{subsec:stage2}
An experimental design DSS with stage $2$ capabilities can monitor operations and issue alerts related to critical experiments throughout the adaptive experimental design process.
For example, alerts may be generated for experiment inputs that: 1) do not align with the experimenter's preferences or risk tolerances based on previous decisions and experiments; 2) lack a sufficient number of observations but may be valuable to conduct; or 3) have an exceedingly high probability of failure, which is an ever-present challenge in field robotics due to the unstructured operational environment. 
As a result, this stage will likely require data related to AUT performance metrics, information gain, and sources of experiment failure. 
These alerts serve to direct the experimenter's attention to some piece of information that could affect the human's decision, but the DSS offers no specific experiment recommendation or remedy for resolving the alert. 
In the literature this kind of support has already demonstrated value in many different applications \cite{wiczorek2014benefits,shishkin2019intelligent,al2020generating}.
The automatically generated alerts in experimental design may be provided to the experimenter using a digital interface (e.g., graphical user interface) and how the experimenter responds to the alerts can provide feedback to the DSS from which its database is updated and decision support is refined.
Alerts are expected to be presented before or during the next experiment so informed decisions can be made.
The specific alert conditions, timing, interface, and information displayed should be reconfigurable to the specific experimenter so that the human does not become complacent and their decision-making is not hindered, as cautioned in the literature \cite{wright2018human}. 

\subsection{Stage 3: Conditional Design Recommendation} \label{subsec:stage3}
This is the first stage for which an experimental design DSS can make recommendations, albeit only under certain experimental conditions.
These conditions fall under two categories: 1) new experiments (i.e., an observation does not yet exist) for which the DSS can only make recommendations for subsystems or components and not specific parameters; or 2) experiments that have already been conducted (i.e., there exists an observation) for which the DSS can only make recommendations to parameters.
The two categories capture the exploration-exploitation tradeoff that experimenters often face when testing field robotic systems; new experiments may reveal capabilities or limitations of the AUT while tuning parameters may improve performance to an acceptable threshold. 
For either scenario, a stage $3$ DSS suggests partially-defined recommendations using a digital interface to define some experiment inputs and the experimenter is responsible for constructing the full experiment definition.
Given the complexity of formulating an intelligent experiment, there are a myriad of opportunities to leverage AI and, as a result, we believe IDSSs will be necessary to realize useful decision support in stage $3$.
AI-enabled capabilities could include searching the design space to optimize parameters of interest, pattern recognition to identify components of interest, and performance inference for information-theoretic experimental design and cost-benefit analysis. 
The quality of decision support will likely be improved by reasoning over the history of decisions made by the experimenter coupled with the observed outcomes.

\subsection{Stage 4: Single Design Recommendation} \label{subsec:stage4}
This is the first stage for which an experimental design DSS is required to make fully-defined recommendations that an experimenter could feasibly execute without modification.
A recommendation from a stage $4$ DSS will define all of the component selections and parameter values that compose the experiment.
The recommendations will also account for the experimenter's preferences and risk tolerance, which may be manually inputted, learned in an automated fashion, and/or inferred from observations of the AUT in previous experiments and interactions with the experimenter.
The DSS will use the experimenter's decision of experimental inputs to construct features, identify trends, and reason about the quantitative results and the experimenter's qualitative assessment within the human-provided context.
Under these conditions, the experimenter is responsible for the decision to accept or reject a recommendation from the DSS.
In the case of rejection, the experimenter is also responsible for defining their choice of experiment input, which could be a partial or full configuration change depending on how much of the recommendation the experimenter chooses to use. 
Finding the optimal decision for recommendation is a heavily-researched topic for a variety of other applications \cite{salama2018decision,hassan2017decision,liu2021clinical}, from which relevant approaches may emerge.

\subsection{Stage 5: Sequential Design Recommendation} \label{subsec:stage5}
A DSS with stage $5$ features can recommend an ordered sequence of fully-defined experiment inputs, all of which take into account the experimenter's preferences and risk tolerance and do not necessarily require human interaction. 
We envision that a sequence of recommendations is the most technologically advanced stage of DSS but could be hugely beneficial for the decision-making process and outcome where the decision maker needs to make sequential decisions over a period of time. 
This is comparable to the long-term planning capability of envisioned DSSs as opposed to the existing DSSs that use short-term planning \cite{zhai2020decision}.

In this stage, the experimenter has no explicit responsibilities but can optionally participate in experimental design decisions by accepting or rejecting the recommendation, which could be any amount of the sequence ranging from a small change in a single experiment to the entire sequence of experiments.
Regardless, conducting the sequence of experiments remains the experimenter's responsibility, as the DSS is only providing a sequence of experiment input definitions.
The anticipated value for defining such a sequence of experiments is for domains where reliable communication is not guaranteed, which is a common challenge in many field robotic applications (e.g., humanitarian assistance and disaster relief, military operations, and space exploration).
Designs for sequences of experiments can serve either as a necessary means of experimentation or a contingency plan to overcome intermittent communications.
In these cases, degraded communication can negatively impact the experimenter's ability to make observations in real-time and configure the AUT for the next experiment.
Examples of this type of experiment condition include long-distance experimentation where the experimenter and AUT are not always physically co-located, such as space robotics where communication to the AUT can take on the order of days.
%
%

\section{Future Directions} \label{sec:future-directions}

In the pursuit of DSSs and IDSSs for experimental design in field robotics, there are a number of open questions and technical gaps that will need to be addressed before a range of decision support can be offered to an experimenter.
We highlight several noteworthy examples here to help direct the scientific community and expect that new challenges will be revealed during future research efforts.

\textbf{System Models.} One pre-requisite for providing principled decision support is a mathematical framework to represent the AUT with respect to the various subsystems, their components, intra-system interactions, and empirical observations.
Such a framework could supplement the existing decision making process with formal methods and information-theoretic approaches.
Importantly, methodologies and representations will need to scale in order to provide decision support for large and complex systems because field robotic systems oftentimes have many subsystems, components, parameters, and permutations of feasible configurations; a probabilistic framework will likely provide the basis for stages $2$ through $5$.
Recent literature on how roboticists can select sensors \cite{rahmani2021sensor} and explore trade-offs in the design space when building robots \cite{saberifar2022charting} may provide useful insights into understanding and modeling systems.

\textbf{Information Models.} In addition to modeling the AUT, future research will be needed to mathematically model information specific to an experimenter, experimental design, and field robotic application. 
The goal of experimentation is to maximize the information gained by the experimenter in the context of their defined objectives while simultaneously reducing the cost to obtain such information.
Interestingly, experimenters typically seek to reduce the number of experiments by avoiding system configurations that result in catastrophic failures; however, these negative examples can be useful to prove infeasibility under real conditions or to generate new, important ideas that robustify autonomy in the AUT.
New models for experimental design, including the value of information and the value of experiments, will be needed to support the decisions that directly affect these competing objectives, comparable to what is referred to in the literature as information economics~\cite{birchler1999information}.
To this end, DSS development would benefit from defining measures of decision effectiveness~\cite{ginzberg1982decision} so that the impact of information and a proposed DSS can be appropriately assessed.
DSSs and IDSSs typically require an appreciable amount of data to contribute intelligent and efficient support, but the cost of conducting field experiments typically limits the number of available observations.
As a result, DSSs and IDSSs will need to be data efficient, include real-time updating, and transform uncertain and incomplete data~\cite{shim2002past} by reasoning effectively with sparse information and inferring the effects of system configurations in order to achieve a more favorable cost-benefit ratio.
%

\textbf{Experimenter Models.} An important aspect of DSS operations is decision support customized to the experimenter.
Building on information-theoretic approaches for the experimenter, future research efforts should consider intuitive and informative methods for modeling, learning, inferring, and incorporating experimenter-specific preferences and risk tolerance.
HRI literature has already investigated a number of human factors pertaining to preferences and risk tolerance, but the sequential decision making process in field experimental design presents unique challenges that require further investigation.
This includes preferences and risk tolerance with respect to a range of different factors (e.g., the experimenter, robotic system, performance, limitations including failure, and experimental costs), as well as dynamic conditions, preferences, and risk tolerance that change based on experimental environments, time constraints, current knowledge, and expected information gain.

\textbf{Decision Making Models.} Planning and decision making across varying time horizons is an open question in virtually all DSS applications, but is especially important in experimental design for field robotics.
A study of a robotic system consists of a number of experiments, all of which work toward the common goal of maximizing the human's understanding of system performance and limitations.
Approaches to orchestrate the sequential decision making of experimental design are required to realize a stage $5$ DSS and could provide useful insights in the maturation of lower stage DSSs to maximize the efficiency of experimentation.
However, this is a particularly difficult problem because it requires inference over both the experimental outcomes with respect to a single, selected experiment configuration and the trend of information gain over multiple experiments with respect to an experimenter.
%

\section{Conclusions} \label{sec:conclusions}
In this work, we propose a six-stage taxonomy of human-centered DSS for adaptive experimental design in the context of field robotics.
We provide supporting terms, definitions, and related work to arrive at a taxonomy that we believe is generally useful for field robotics experimentation and provides a common framework for future research.
To provide more assistance to the research community, we also identify technical gaps and specific future directions with respect to the requirements in our proposed taxonomy and the existing literature.


\bibliography{main}

\end{document}